\documentclass[sigconf,screen,authorversion,nonacm,natbib=false]{acmart}

\usepackage[utf8]{inputenc}
\usepackage[style=ACM-Reference-Format,backend=biber,defernumbers=false,sortcites=true]{biblatex}
\usepackage{subcaption}
\usepackage{multirow}
\usepackage{xspace}
\usepackage{listings}
\usepackage{booktabs}

\RequirePackage{fontawesome}

\usepackage{xcolor, soul}
\sethlcolor{yellow}

\newcommand{\reddit}[2]{
    \href{https://www.reddit.com/r/GamePhysics/comments/#1}{{\smaller\faExternalLink}\xspace#2}%
}

\begin{document}

    \title{CLIP meets GamePhysics: Towards bug identification in gameplay videos using zero-shot transfer learning}
    \author{Mohammad Reza Taesiri}
\email{taesiri@ualberta.ca}
\affiliation{
  \institution{University of Alberta}
  \city{Edmonton}
  \state{AB}
  \country{Canada}
}
\author{Finlay Macklon}
\email{macklon@ualberta.ca}
\affiliation{
  \institution{University of Alberta}
  \city{Edmonton}
  \state{AB}
  \country{Canada}
}
\author{Cor-Paul Bezemer}
\email{bezemer@ualberta.ca}
\affiliation{
  \institution{University of Alberta}
  \city{Edmonton}
  \state{AB}
  \country{Canada}
}
    \date{January 2022}
	
    \begin{abstract}

Gameplay videos contain rich information about how players interact with the game and how the game responds.  Sharing gameplay videos on social media platforms, such as Reddit, has become a common practice for many players. Often, players will share gameplay videos that showcase video game bugs. Such gameplay videos are software artifacts that can be utilized for game testing, as they provide insight for bug analysis. Although large repositories of gameplay videos exist, parsing and mining them in an effective and structured fashion has still remained a big challenge. 
In this paper, we propose a search method that accepts any English text query as input to retrieve relevant videos from large repositories of gameplay videos. Our approach does not rely on any external information (such as video metadata); it works solely based on the content of the video. By leveraging the zero-shot transfer capabilities of the Contrastive Language-Image Pre-Training (CLIP) model, our approach does not require any data labeling or training. To evaluate our approach, we present the \texttt{GamePhysics} dataset consisting of 26,954 videos from 1,873 games, that were collected from the GamePhysics section on the Reddit website. Our approach shows promising results in our extensive analysis of simple queries, compound queries, and bug queries, indicating that our approach is useful for object and event detection in gameplay videos. An example application of our approach is as a gameplay video search engine to aid in reproducing video game bugs. Please visit the following link for the code and the data: \href{https://asgaardlab.github.io/CLIPxGamePhysics/}{https://asgaardlab.github.io/CLIPxGamePhysics/}


	\end{abstract}
	
    

    \settopmatter{printfolios=true}
	\maketitle

    \section{Introduction} \label{sec:intro}

Video game development is a highly complex process.
There are many unique challenges when applying general software engineering practices in video game development~\cite{politowski2020dataset, murphy2014cowboys, stacey2009temporal, petrillo2009went, lewis2010went}, including challenges in game testing.
Manual testing is a widely accepted approach to game testing~\cite{politowski2021survey, varvaressos2017automated, pascarella2018video}, however this manual process is slow and error-prone, and most importantly, expensive.  
On the other hand, it is challenging to automate game testing~\cite{lewis2011whats, pascarella2018video, santos2018computer} due to the unpredictable outputs of video games.
Despite progress towards automated game testing methods~\cite{davarmanesh2020automating, ling2020using, varvaressos2017automated, tuovenen2019mauto} and game testing tools~\cite{juliani2018unity, zheng2019wuji, bergdahl2020augmenting, pfau2020dungeons}, new approaches to game testing must be researched. 

The difficulty of game testing due to the unique nature of games calls for unique testing methodologies as well. 
For example, we could leverage the visual aspect of games in the testing process.
Having a gameplay video is very helpful when trying to reproduce a bug in the development environment for further analysis, as bug reports often contain incomplete information~\cite{bettenburg2008makes}.
The ability to search a large repository of gameplay videos with a natural language query would be useful to help reproduce such bug reports.
For example, in the game development domain, a bug report might state `my car is stuck on the rooftop' without any screenshot or video to show what is actually happening. 
A gameplay video search would allow game developers to find an example instance of a specific bug in the pile of gameplay videos from their playtesting sessions or the internet (e.g., YouTube, Twitch). 

Despite containing rich information, the challenges related to video parsing and understanding mean that gameplay videos are difficult to utilize.
Manually identifying bug instances is time consuming, and there is limited prior research on automatic methods for mining large repositories of gameplay videos~\cite{lin2019identifying, luo2019making}.

\begin{figure}[!t]
    \centering
    \includegraphics[width=\columnwidth]{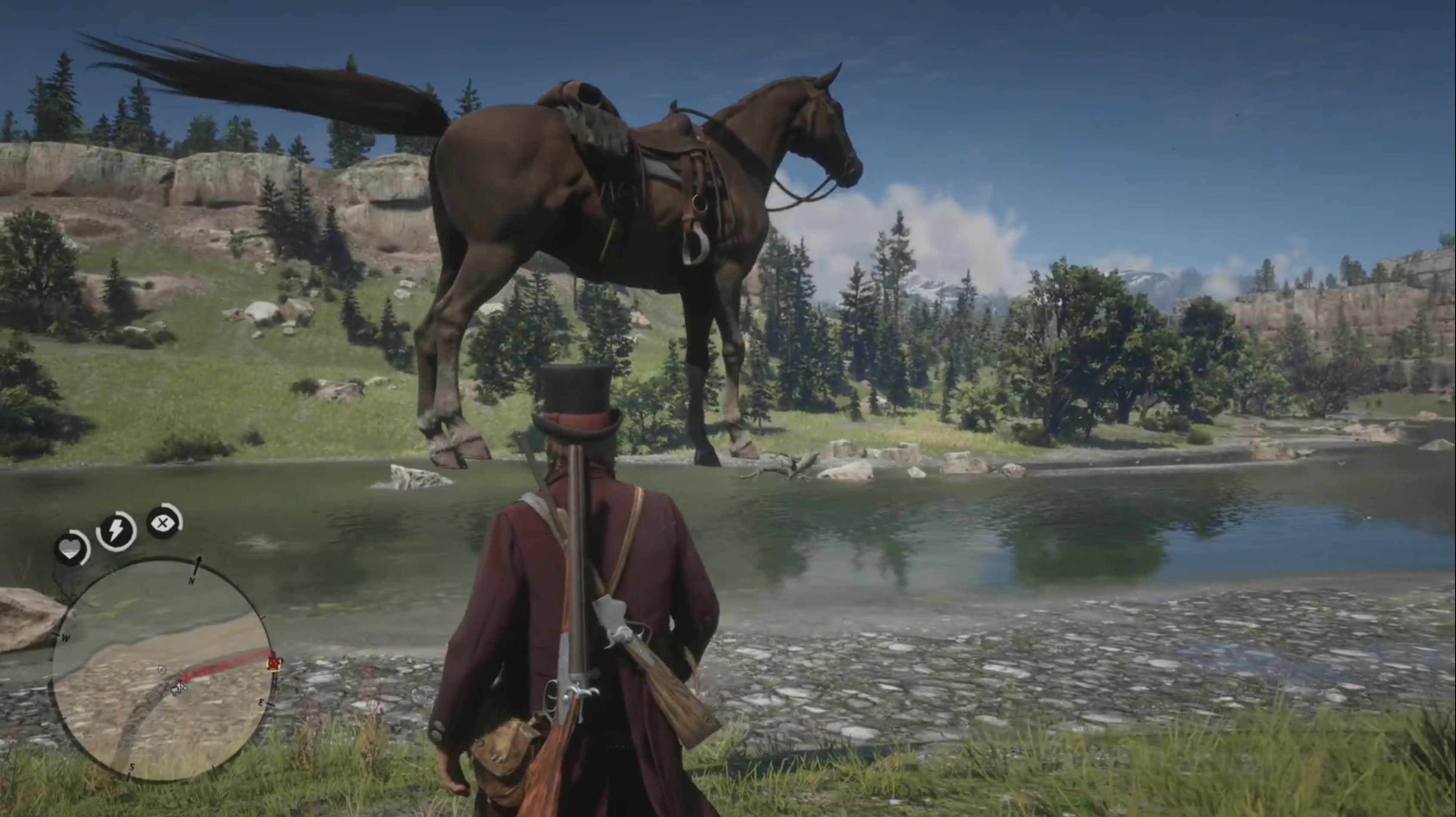}
    \caption{\reddit{9rqabp}{Video} identified by our approach with the bug query \textit{`A horse in the air'} for Red~Dead~Redemption~2.}
    \label{fig:search_example}
\end{figure}

In this paper, we address the challenges of extracting useful information from large repositories of gameplay videos.
We propose an approach for mining gameplay videos using natural language queries by leveraging the Contrastive Language-Image Pre-Training (CLIP) model~\cite{radford2021learning} to identify similar text-image pairs without any additional training (i.e., zero-shot prediction). We leverage CLIP for videos by pre-processing the video frame embeddings and use Faiss~\cite{JDH17} to perform a fast similarity search for the pairs of text queries and video frames.
In our approach, we present two methods to aggregate across the similarity scores of each text-image pair to identify relevant videos.
To evaluate our approach, we collected and prepared the \texttt{GamePhysics} dataset, consisting of 26,954 gameplay videos that predominantly contain game physics bugs.
We evaluate our approach with sets of simple queries, compound queries, and bug queries, and show that our approach can identify objects and (bug-related) events in large repositories of gameplay videos.
Figure~\ref{fig:search_example} shows an example of a video that was identified by our approach when searching videos from the Red Dead Redemption~2 game using the bug query \textit{`A horse in the air'}.
The primary application of our approach is as a gameplay video search engine to aid in reproducing game bugs.
With further work, e.g. setting thresholds to limit false positives, our approach could also be used as a bug detection system for video games.

The main contributions of our paper are as follows:
\begin{itemize}
    \item We propose an approach to search for objects and events in gameplay videos using a natural language text query.
    \item We collect and prepare the \texttt{GamePhysics} dataset, consisting of 26,954 gameplay videos from 1,873 games. 
    \item We report results that demonstrate the promising performance of our approach in identifying game physics bugs through 3 experiments with simple, compound, and bug queries.
\end{itemize}

The remainder of our paper is structured as follows. In Section~\ref{sec:background}, we motivate our study by providing relevant background information. In Section~\ref{sec:related}, we discuss related work. In Section~\ref{sec:approach}, we present our approach to mining large repositories of gameplay videos. In Section~\ref{sec:dataset}, we discuss collecting and pre-processing the \texttt{GamePhysics} dataset. In Section~\ref{sec:experiments} we detail our experiment setup, and in Section~\ref{sec:results} we present our results. In Section~\ref{sec:discussion} we provide discussion and insights on the performance of our approach. In Section~\ref{sec:limitations} we outline limitations of our approach. In Section~\ref{sec:threats} we address threats to validity. We conclude our paper in Section~\ref{sec:conclusion}.

 	\section{Motivation and Background} \label{sec:background}

\begin{figure*}[!t]
	\centering
	\begin{subfigure}[t]{\columnwidth}
		\centering
		\includegraphics[width=\textwidth]{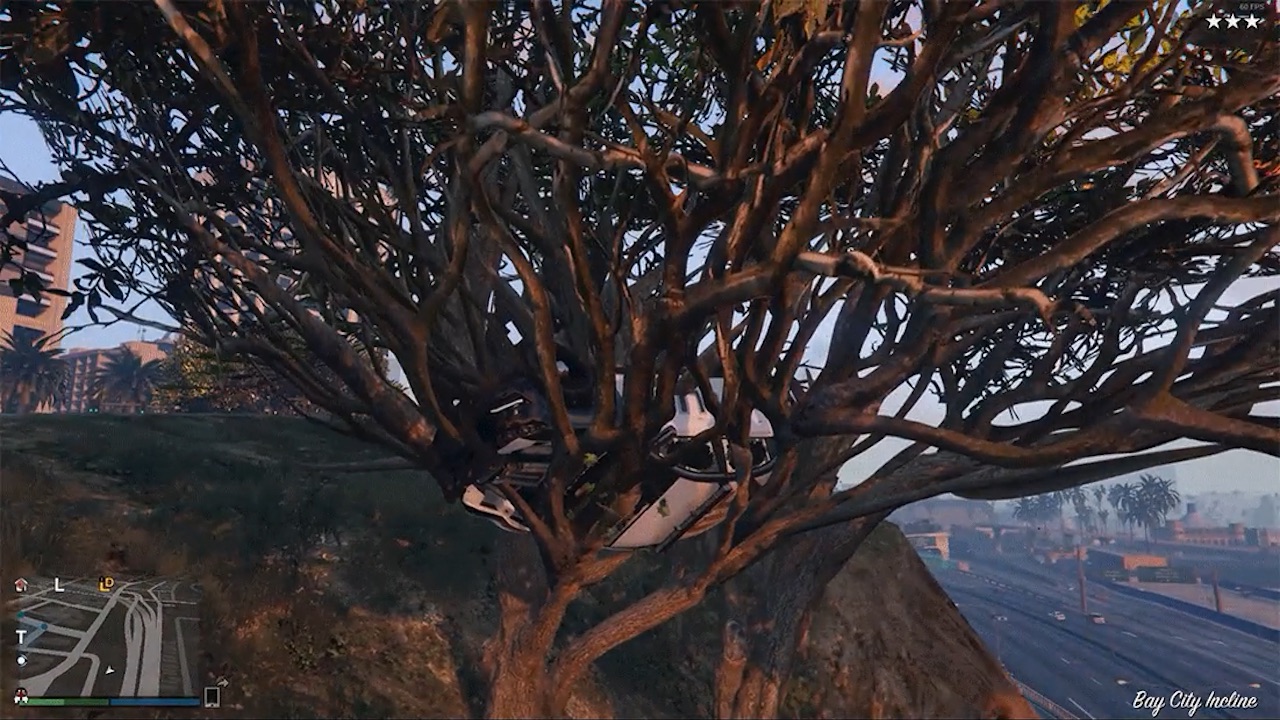}
		\caption{\reddit{4jirzj}{Bug} in Grand Theft Auto~V. Car stuck in a tree after colliding.}
		\Description{Fully described in the text.}
		\label{fig:sample_bug_1}
	\end{subfigure}
	\hfill
	\begin{subfigure}[t]{\columnwidth}
		\centering
		\includegraphics[width=\textwidth]{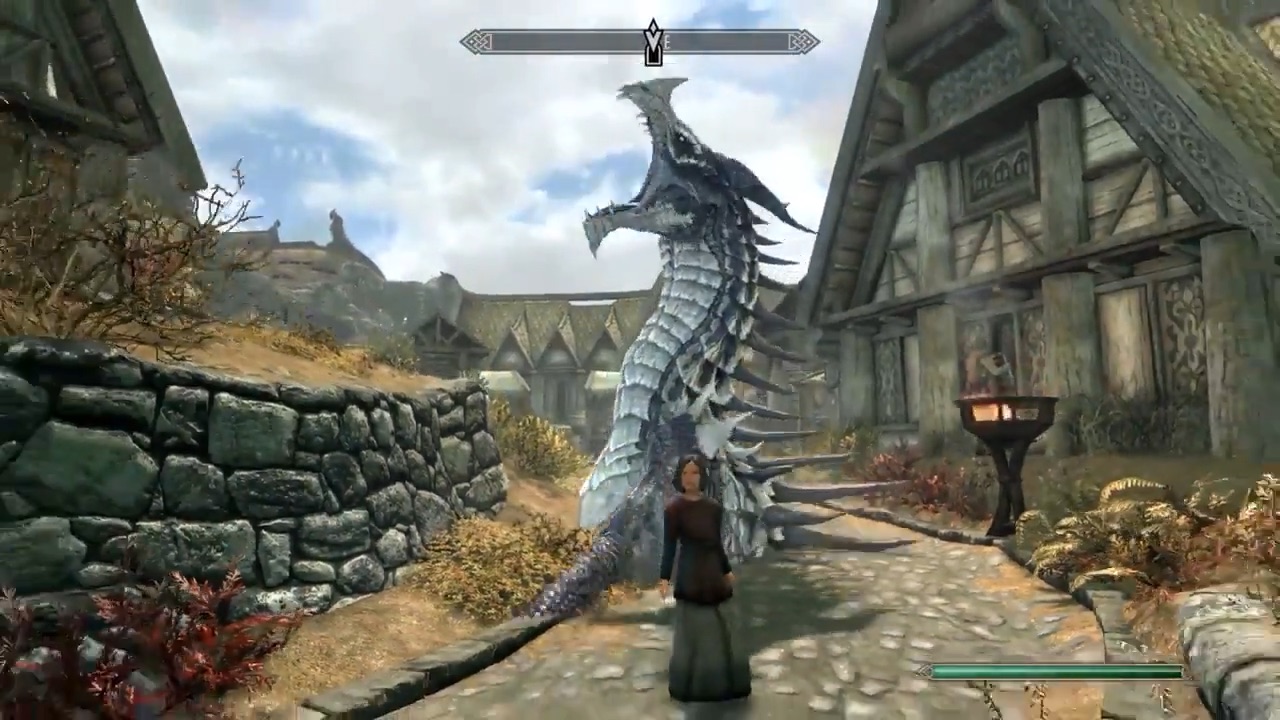}
		\caption{\reddit{6652mm}{Bug} in The Elder Scrolls~V: Skyrim. Dragon stuck in the ground.}
		\Description{Fully described in the text.}
		\label{fig:sample_bug_2}
	\end{subfigure}
	\hfill
	\begin{subfigure}[t]{\columnwidth}
		\centering
		\includegraphics[width=\textwidth]{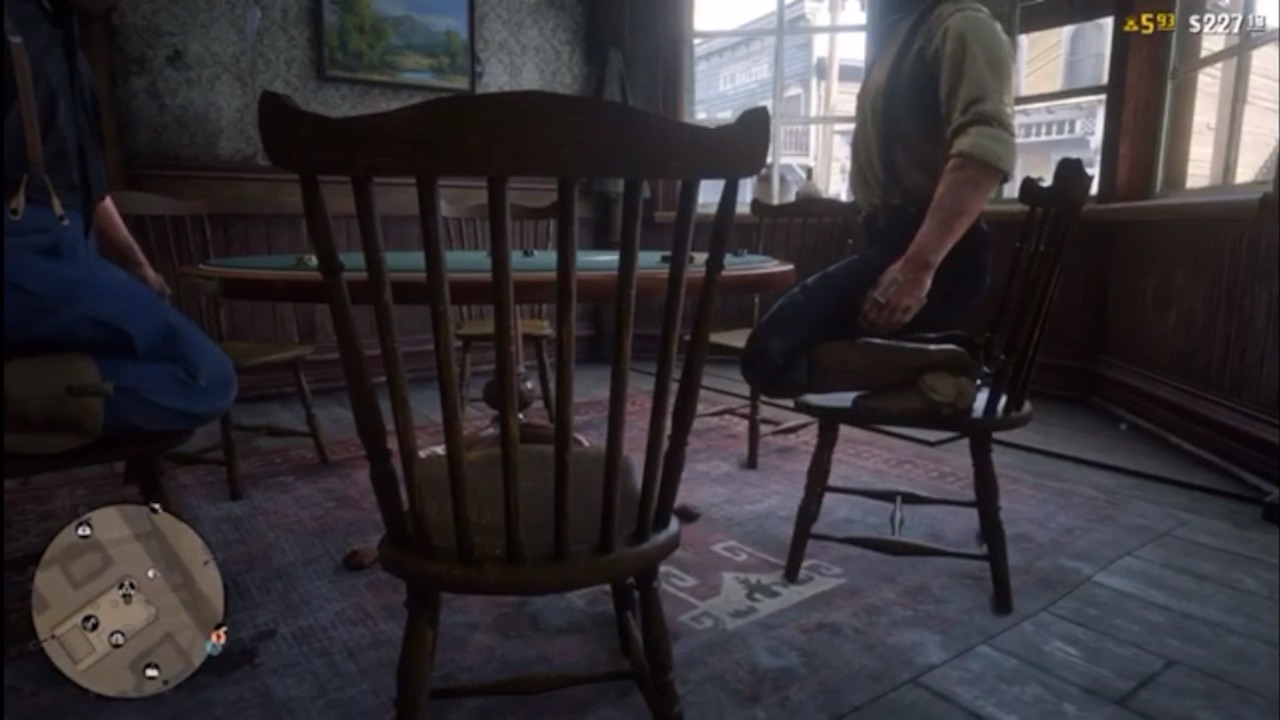}
		\caption{\reddit{bur1ke}{Bug} in Red Dead Redemption~2. Incorrect sitting animation.}
		\Description{Fully described in the text.}
		\label{fig:sample_bug_3}
	\end{subfigure}
	\hfill
	\begin{subfigure}[t]{\columnwidth}
		\centering
		\includegraphics[width=\textwidth]{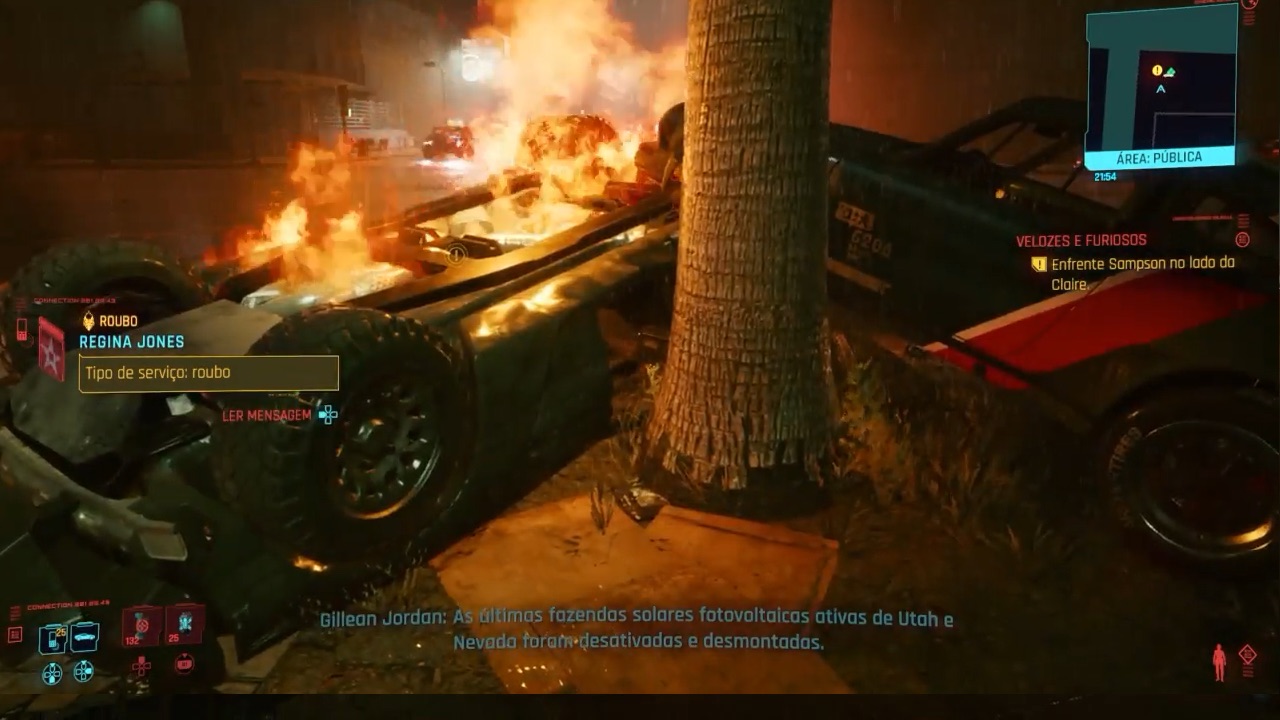}
		\caption{\reddit{kv41nk}{Bug} in Cyberpunk~2077. Cars stuck together after colliding.}
		\Description{Fully described in the text.}
		\label{fig:sample_bug_4}
	\end{subfigure}
	\caption{Sample instances of game physics bugs.}
	\Description{Fully described in the text.}
	\label{fig:sample_bugs}
\end{figure*}

\subsection{Video game (physics) bugs} \label{subsec:bugs}

In this paper, we are interested in a specific category of bugs in video games that we call `game physics' bugs. 
Game physics bugs are not necessarily related to an inaccurate physics simulation.
Many of these bugs are related to the faulty representation of game objects due to an error in the internal state of that object.
A few sample instances of game physics bugs can be seen in Figure~\ref{fig:sample_bugs}.
In Figure~\ref{fig:sample_bug_1}, a bug from Grand Theft Auto~V related to object collisions is shown.
Figure~\ref{fig:sample_bug_2} shows a bug from The Elder Scrolls~V: Skyrim, related to object clipping.
In Figure~\ref{fig:sample_bug_3}, a bug from Red Dead Redemption~2 related to ragdoll poses can be seen.
Figure~\ref{fig:sample_bug_4} shows a bug from Cyberpunk~2077, related to object collisions.
Identifying game physics bugs is challenging because we need to be able to extract specific, high-level (abstract) events from the gameplay videos, that are often similar to correct behaviour. 


\subsection{Challenges in mining gameplay videos} \label{subsec:challenges}
Until now, it has been challenging to extract valuable information from large repositories of gameplay videos.
Identifying bug instances by manually checking the contents of gameplay videos is time-consuming ~\cite{lin2019identifying}.
Therefore, automatic methods for mining gameplay videos are required.
The only existing approach for automatically extracting events from gameplay videos requires manual data labelling (and the training of new models)~\cite{luo2019making}, which itself is time-consuming.
Therefore, an effective method for extracting valuable information from gameplay videos should be able to automatically analyze the video contents without requiring manual data labelling.

\subsection{Contrastive learning and zero-shot transfer} \label{subsec:contrastive}
While there are many approaches towards zero-shot learning, we are interested in assessing the zero-shot performance of pre-trained contrastive models. 
Contrastive learning is a machine learning technique in which the goal is to learn a representation of inputs such that similar items stay close to each other in the learned space, while the dissimilar items are far away~\cite{becker1992self, bromley1993signature}.
In recent years, contrastive learning has been one of the key drivers in the success of self-supervised learning methods and has been used for zero-shot transfer learning~\cite{chen2020simple, DBLP:conf/nips/GrillSATRBDPGAP20, khosla2020supervised, radford2021learning}. 
Zero-shot learning is a family of problems in machine learning, in which an algorithm is required to solve a task without having a training set for that specific task 
~\cite{5206594, larochelle2008zero}. 
To illustrate this idea, suppose that a person has never seen a zebra before. 
If we give a detailed description of a zebra to them (e.g., an animal similar to a horse, but with black-and-white stripes all over their bodies), that person can identify a zebra when they see one.

\subsection{The Contrastive Language-Image Pre-Training (CLIP) model} \label{subsec:clip}
One contrastive model that has proven zero-shot transfer capabilities is the Contrastive Language-Image Pre-Training (CLIP) model~\cite{radford2021learning}, which can leverage both text and image inputs together. We decided to use CLIP because of its multimodal capabilities and the size of its training dataset. The CLIP model consists of two parts: a text encoder, and an image encoder. 
These two parts work individually, and they can accept any English text and image as input. 
When an encoder of this model receives an input, it will transform it into an embedding vector. 
These embedding vectors are high-level features that are extracted by the network, representing the input. 
More specifically, these embedding vectors are how the neural network represents, distinguishes, and reasons about different inputs. 
Both encoders of this model will produce vectors of the same dimension for image and text inputs. 
Not only do these vectors have the same dimension, but they are also in the same high-dimensional feature space, and are therefore compatible with each other. 
For example, the embedding vector of the text `an apple' and the embedding vector of an image of an apple are very close to each other in this learned space.
The CLIP model was pre-trained on over 400 million pairs of images and text descriptions that were scraped from the internet, and has six different backbone architectures: `RN50', `RN101', `RN50x4', `RN50x16', `ViT-B/32', `ViT-B/16'. 
The models with `RN' in their name are ResNet-based~\cite{he2016deep} models using traditional convolutional layers, while the `ViT' models are based on vision transformers~\cite{dosovitskiy2020image}.

 	\section{Related Work} \label{sec:related}
Event extraction from video content is of special importance for various data mining tasks~\cite{macleod2015code, ponzanelli2016too}. Only two prior studies have explicitly explored automatic approaches for mining gameplay videos, with varying success. Lin et al. showed that using metadata (such as keywords) to identify YouTube videos that contain video game bugs is feasible~\cite{lin2019identifying}, but our approach looks at the video contents, which Lin et al. do not take into account.
Our approach is more useful for game developers, as we can identify objects and (bug-related) events within gameplay videos.
Luo et al. propose an approach for automatic event retrieval in e-sport gameplay videos that requires manual data labelling, a fixed set of classes (events), and the training of new models~\cite{luo2019making}.
Our approach is more robust and easier to set-up, as we can search gameplay videos with any English text query to identify specific objects and events without performing manual data-labelling.

Although there is limited prior work on mining large repositories of gameplay videos, there are several studies that propose approaches to automatically detect graphics defects in video games.
One of the earliest approaches for automated detection of graphics defects was published in 2008, in which a semi-automated framework was proposed to detect shadow glitches in a video game using traditional computer vision techniques~\cite{nantes2008framework}. 
Recent studies have utilized convolutional neural networks in their approach to automatically detect a range of graphics defects~\cite{davarmanesh2020automating, ling2020using, taesirivideo, chen2021glib}.
Instead of detecting graphics defects, our work is concerned with the automatic identification of game physics bugs in gameplay videos.

Tuovenen et al. leverage the visual aspect of games through an image matching approach to create a record-and-replay tool for mobile game testing~\cite{tuovenen2019mauto}.
Our approach leverages the visual aspect of games in a different way; instead of recording tests through gameplay, we automatically identify bugs in gameplay videos.

Some studies have proposed approaches for automated detection of video game bugs through static or dynamic analysis of source code.
Varvaressos et al. propose an approach for runtime monitoring of video games, in which they instrument the source code of games to extract game events and detect injected bugs~\cite{varvaressos2017automated}.
Borrelli et al. propose an approach to detect several types of video game-specific bad smells, which they formalize into a tool for code linting~\cite{borrelli2020detecting}. 
Our approach differs as we do not require access to the source code of games; instead we identify video game bugs based solely on the contents of gameplay videos.

In addition to related work on automatic bug detection for video games, there exists a wide range of work that leverages recent advancements in deep learning to provide new tools and techniques that address problems faced by game developers.
Several studies have sought to make AI methods accessible in the video game development and testing cycle, either through the game's internal state, raw pixels, or through a high-level neural network-based representation~\cite{khameneh2020embedding, trivedi2021contrastive}. 
Some studies have proposed approaches to accompany a game designer through the creation process of a game by providing suggestions and explanations to the designer~\cite{guzdial2016game, DBLP:conf/aiide/GuzdialLR18, khadivpour2020responsibility}. 
Other studies have incorporated reinforcement learning and evolutionary methods to create AI agents that can automatically play games~\cite{justesen2019deep, berner2019dota, vinyals2019grandmaster}. 
These AI agents can be further employed to perform automated game testing sessions~\cite{roohi2021predicting, 9619048, 9231552, zheng2019wuji, GARCIASANCHEZ2018133}.
Our work is different from those listed above, as we focus on assisting game developers by providing an approach to efficiently search large repositories of gameplay videos to find bug instances.

 	\section{Our Approach} \label{sec:approach}
\begin{figure}[!t]
\centering
\includegraphics[width=\linewidth]{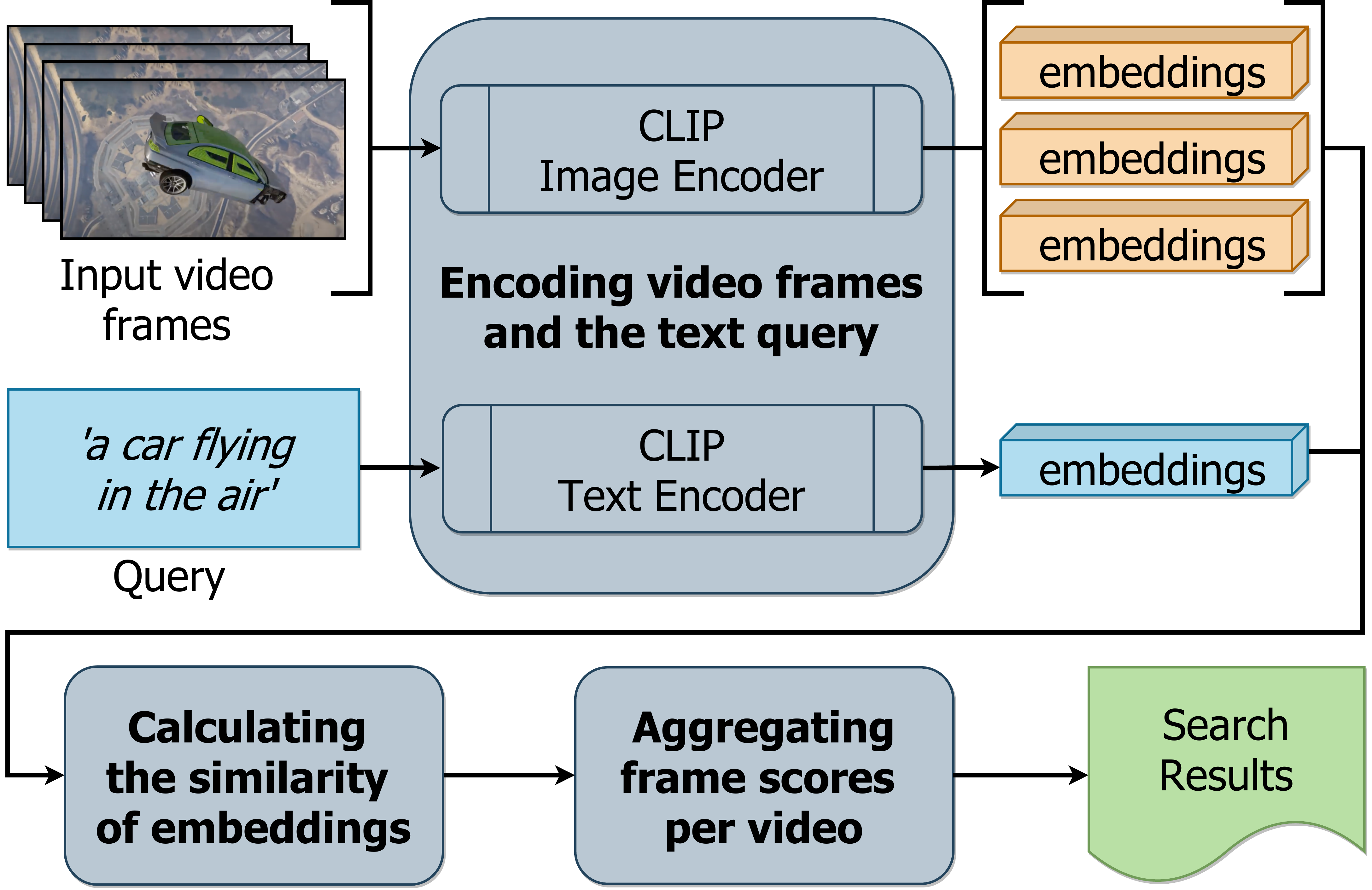}
\caption{Overview of our gameplay video search approach.}
\label{fig:videosearch}
\end{figure}

To assist with detection and analysis of game bugs, we propose an approach that quickly and effectively searches a large repository of gameplay videos to find a specific object or event in a particular game.
For creating such a powerful search system, one could utilize a traditional supervised classification technique. 
However, any supervised classification method requires a training dataset, a test dataset, and a fixed number of classes. 
Maintaining these two datasets and labeling each sample is demanding and labour-intensive.
On the other hand, the CLIP model provides zero-shot transfer learning capabilities that allow us to develop an approach to automatically mine gameplay videos while avoiding the aforementioned issues. 
Figure \ref{fig:videosearch} shows an overview of our approach.

\subsection{Encoding video frames and the text query} \label{subsec:encoding}
Our approach accepts a set of videos and any English text query as inputs.
We first extract all frames from each video, and then use the CLIP model to transform our input text query and input video frames into the embedding vector representations described in Section~\ref{subsec:clip}.
We selected the CLIP model because it is flexible enough to accept any arbitrary English text as a query and compare it with a video frame, without any additional training.

\subsection{Calculating the similarity of embeddings} \label{subsec:similarity}
As well as avoiding manual data labelling, our approach avoids depending upon any extra information, such as metadata, to search gameplay videos.
Instead, we are able to calculate similarity scores solely based on the contents of the video frames and the text query.
The similarity score in our problem is a distance between an embedding vector representing a text query and another embedding vector representing a video frame.
To calculate similarity scores for the pairs of embedding vectors, we opted for cosine similarity, a widely-used similarity metric ~\cite{viggiato2021identifying, 10.1007/978-3-030-41131-2_3, fazli2021under, viggiatousing}. 
We require an exhaustive search to calculate the similarity score of the text query with each individual frame in each input video. 
The performance of an exhaustive search will decrease inversely with an increasing number of videos in a repository. 
To combat this, we use Faiss~\cite{JDH17} to conduct an efficient similarity search.

\subsection{Aggregating frame scores per video} \label{subsec:aggregating}
Although CLIP is designed to accept text and images as inputs, we can leverage CLIP for videos by treating each video as a collection of video frames (i.e. a collection of images). 
To identify specific events that could occur at any moment in a gameplay video, we cannot subsample the video frames as suggested in the original CLIP, because due to the richness of events in a single gameplay video,  skipping any part of the video may lead to information loss and inaccurate results. 
Therefore, we perform a similarity search on all frames of all videos by comparing each individual video frame with the target query text, and we subsequently aggregate the similarity scores across each video.
Below we detail the design of two different methods for aggregating the video frame similarity scores for each gameplay video.
Our approach supports the two aggregation methods without the need to re-calculate the similarity scores. 

\subsubsection*{Aggregating frame scores using the maximum score}
Our first score aggregation method ranks videos based on the maximum similarity score across all frames belonging to each video.
This method is highly sensitive, as a single frame with high similarity can lead to an entire video being identified as relevant to the query.

\subsubsection*{Aggregating frame scores using the similar frame count}
In the second score aggregation method, we begin by ranking all frames of the input videos based on their similarity scores with the text query.
Then, we select a predefined number (the \emph{pool size} hyperparameter) of highest-ranked frames across all videos. 
Finally, we count the number of frames per video within this pool of highest-ranked frames. 
This method is less sensitive than our first aggregation method, as identified videos must have multiple frames that are among the most similar to the input text query.
We selected 1,000 as the default pool size value in our study.


 	\section{Preparing the GamePhysics dataset} \label{sec:dataset}

\subsection{Collecting the \texttt{GamePhysics} dataset} \label{subsec:collecting}

\begin{figure}[!t]
\centering
\includegraphics[width=\linewidth]{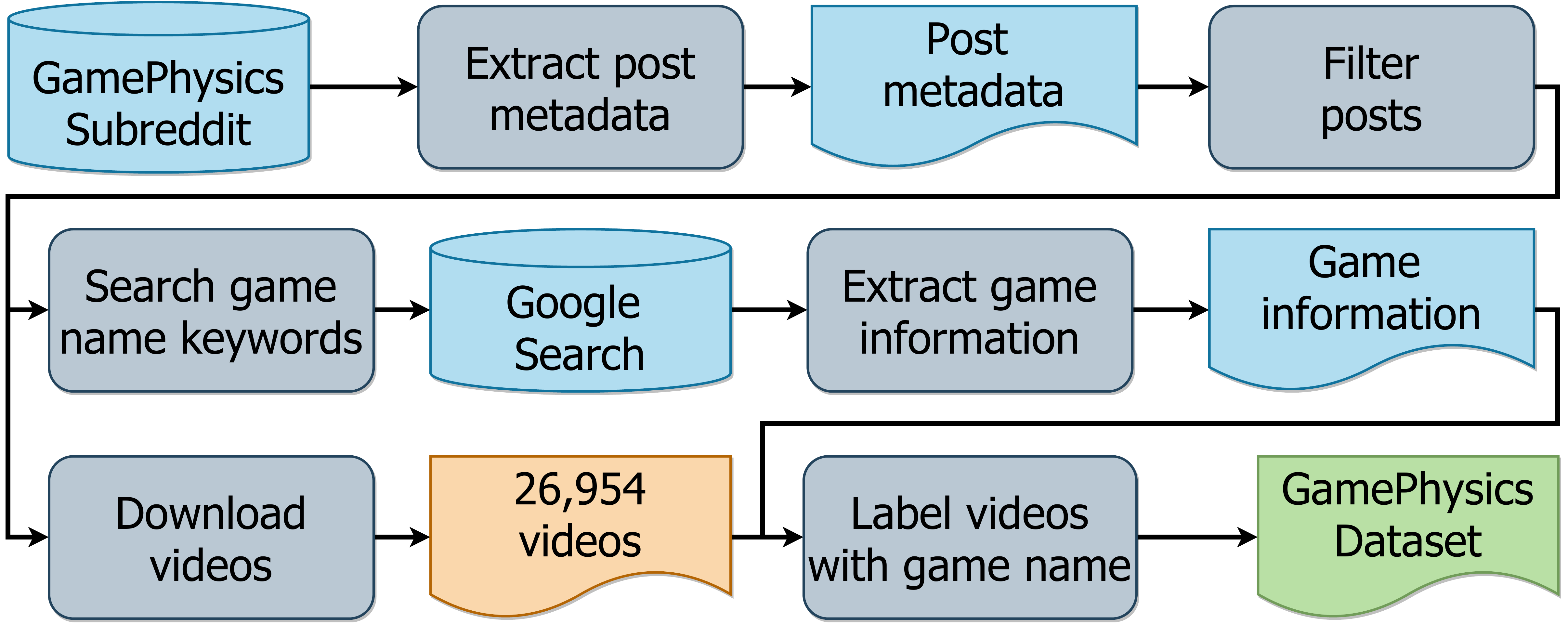}
\caption{Overview of our data collection process.}
\label{fig:datacollection}
\end{figure}

Developing and testing a new machine learning system requires a dataset. Unfortunately, there is no such dataset for gameplay bugs. To this end, we present the \texttt{GamePhysics} dataset, which consists of \textbf{26,954} gameplay videos collected from the \href{https://www.reddit.com/r/GamePhysics/}{{\smaller\faExternalLink} GamePhysics} subreddit. 
An overview of our data collection process can be seen in Figure~\ref{fig:datacollection}.

\subsubsection*{Extracting post metadata and downloading videos}
To collect the data, we created a custom crawler that uses both the official Reddit API and the popular \texttt{PushShift.io} API \cite{baumgartner2020pushshift}. 
In our crawler, we use the \texttt{PushShift.io} API to get high-level information about each submission in the GamePhysics subreddit. 
After obtaining high-level data, we use Reddit's official API to update the scores and other metadata of each submission. 
For downloading the actual video files, we use a combination of \texttt{youtube-dl} and \texttt{aria2c} to extract video links and download them.

\subsubsection*{Filtering posts}
We applied several filters to our dataset during the data collecting process to remove spam posts, low-quality content, and outliers. 
There are several spam posts in the GamePhysics subreddit, and these posts are marked explicitly as spam by the subreddit's moderators. 
Furthermore, we treat post scores as a quality signal as this score captures up/down votes from Reddit users, and consider any post with a score of less than one as low-quality content. 
The lengths of the video files vary from a few seconds to multiple hours. 
We avoid long videos in our dataset, because they can contain multiple events of different kinds and are very hard to process. 
We only keep videos that are longer than 2 seconds and shorter than 60 seconds. 
After applying our filters, our final dataset contains \textbf{26,954} video files from \textbf{1,873} different games.

\subsubsection*{Labelling videos with the game name}
In order to simulate the realistic scenario in which a game developer would search a repository of gameplay videos for a specific game, we extract the game name for each gameplay video from the title of its respective post.
Detecting the game's name from a GamePhysics submission is not straightforward. 
While there is a community guideline that suggests including the name of the game in the submission's title, people often forget to include the game name or use several aliases for the game name, meaning the task of detecting the game name can be hard. 
For example, `GTA V' is a widely-used alias that refers to the `Grand Theft Auto V' game. 
To address this issue, we created a second custom crawler to search game name keywords in Google and subsequently map them to the full game name. 
Google search results provide a specific section called the Knowledge Panel that contains the game name, as well as other relevant game information such initial release date, genre, development studio(s), and publisher.

\subsection{Pre-processing the videos} \label{subsec:preprocessing}
As discussed in Section~\ref{subsec:similarity}, our approach can search a large repository of gameplay videos more efficiently by pre-processing the embedding vectors of every frame for each video in the repository before inputting any text queries. 
Therefore, for our dataset to be suitable for our approach, we pre-process all videos in the \texttt{GamePhysics} dataset before proceeding with any experiments.
We pre-processed all 26,954 videos using a machine with two NVIDIA Titan RTX graphics cards, but it is certainly possible to perform this step with less powerful graphics cards too.
It is worth noting that this is by far the most computationally expensive step in our approach.
 	\section{Experiment setup} \label{sec:experiments}
In this section, we describe an extensive analysis of our approach on the \texttt{GamePhysics} dataset through a diverse set of experiments.
To assess the performance of our video search method, we performed several experiments with varying levels of difficulty. 
The main obstacle to evaluating our search system is the lack of a benchmark dataset. 
To this end, we designed three experiments with three corresponding sets of queries to shed light on the capabilities of our proposed method.

\subsection{Experiment overview} \label{subsec:expoverview}
In the first two experiments, we evaluate the accuracy of our approach when retrieving videos with certain objects in them. 
The results for this step indicate the generalization capability of the model for the third experiment.
In the third experiment, we evaluate the accuracy of our approach when retrieving videos with specific events related to bugs.

\subsection{Selecting CLIP architectures} \label{subsec:selectingclips}
To understand the relative performance of the available ResNet-based and vision transformer-based CLIP models, we opted to try two different backbone architectures in our system, namely `RN101' and `ViT-B/32'. 
We chose these backbones as fair baseline comparisons because they are the largest backbone architectures in their respective families, assuming we stipulate equivalent input image sizes ($224\times224$).
For comparison, the `ViT-B/32` backbone architecture contains 151,277,313 total parameters, while the `RN101` backbone architecture contains 119,688,033 total parameters.
We selected the largest architectures as we are performing inference with these models, not training them.

\begin{table*}[!t]
    \centering
    \caption{Games selected for evaluation of our approach. All selected games are open-world.}
    \begin{tabular*}{\textwidth}{l @{\extracolsep{\fill}} l @{\extracolsep{\fill}} l @{\extracolsep{\fill}} l @{\extracolsep{\fill}} l @{\extracolsep{\fill}} r}
        \toprule
        \textbf{Game} & \textbf{Key} & \textbf{Genre} & \textbf{Visual style} & \textbf{Reason for inclusion} & \textbf{Videos} \\
        \midrule
        Grand Theft Auto V         & \texttt{GTA} & Action-adventure & Realism & Variety of vehicles & 2,230 \\
        Red Dead Redemption 2      & \texttt{RDR} & Action-adventure & Realism & Historical style & 754 \\
        Just Cause 3               & \texttt{JC3} & Action-adventure & Realism & Physical interactions & 680 \\
        Fallout 4                  & \texttt{F4} & Action role-playing-game & Fantasy realism (Retro-futuristic) & Unique look and feel & 614 \\
        Far Cry 5                  & \texttt{FC5} & First-person shooter & Realism & First-person camera & 527 \\
        Cyberpunk 2077             & \texttt{C77} & Action-adventure & Fantasy realism (Futuristic) & High-quality lighting & 511 \\
        The Elder Scrolls V: Skyrim & \texttt{ESV} & Action role-playing-game & Fantasy realism & Magical effects & 489 \\
        The Witcher 3: Wild Hunt   & \texttt{W3} & Action role-playing-game & Fantasy realism & Mythical beasts & 387 \\
        \bottomrule
    \end{tabular*}
    \label{tab:selected_games}
\end{table*}

\subsection{Selecting video games} \label{subsec:selectinggames}
Our dataset contains videos from 1,873 different video games, and the differences in their high-level characteristics, such as genre, visual style, game mechanics, and camera view, can be vast. 
Therefore, we performed a comprehensive evaluation in all three experiments with 8 popular video games that differ in their high-level characteristics.
The only uniting characteristic for our selected games is that they have open-world mechanics, because developers of open-world games would find particular benefit from an effective video search for bug reproduction.
Open-world games allow a player to freely explore the game world, providing a larger set of potential interactions between the player and game environment.
Open-world games are therefore more likely to suffer from game physics bugs that are difficult to reproduce.
Table~\ref{tab:selected_games} shows each game we selected for our experiments, as well as some game characteristics and the reason for inclusion.
In total, 23\% of videos in the \texttt{GamePhysics} dataset are from these 8 video games (6,192 videos).

\subsection{Query formulation} \label{subsec:formulation}
To come up with a set of relevant search queries in the experiments, we randomly picked 10 videos from each of the 8 selected games. 
The first author manually reviewed each of the 80 samples to understand what was happening and how we could describe events in gameplay videos that contain bugs.
This sampling process helped us pick relevant objects and events to use in our queries.

\subsection{Experiment 1: Simple Queries} \label{subsec:exp1}
In this experiment, we searched for specific objects in videos, e.g. a car.
Our main objective in this experiment is to demonstrate the capability of our system for effective zero-shot object identification. As a reminder, we never trained or fine-tuned our neural network model for any of these experiments or any video game.  
We created 22 distinct queries for Experiment~1, including transportation vehicles, animals, and special words describing the weather or environment. 
For this experiment we wanted our approach to operate with very high sensitivity, and so we selected our first aggregation method, i.e. using maximum frame score per video (Section~\ref{subsec:aggregating}).

\subsection{Experiment 2: Compound Queries} \label{subsec:exp2}
Continuing our evaluation, we search for compound queries, i.e. queries in which an object is paired with some descriptor.
Similar to Experiment~1, we only use compound queries that are relevant to each video game.
For example, in the previous experiment, we searched for videos in the Grand Theft Auto~V game that contained a car, but in this experiment we evaluate the performance of our approach when searching for objects with a specific condition, like a car with a particular color. 
For this second experiment, we created a set of 22 compound queries, and again selected our first aggregation method (using maximum frame score per video).

\subsection{Experiment 3: Bug Queries} \label{subsec:exp3}
In the third experiment, we search for bug queries, i.e. phrases that describe an event in the game which is related to a bug. 
We manually created specific textual descriptions for different bug behaviors in the game and searched our dataset to see if we could detect bugs in gameplay videos. 
Similar to the previous experiments, our bug queries are game-specific. 
For this experiment, we created a set of 44 unique bug queries, with each query describing an event.
Given the complex nature of the bug queries in Experiment~3, we decided to use our less sensitive aggregation method, based on the number of highly similar frames per video (as described in Section~\ref{subsec:aggregating}).

\subsection{Evaluating the experiments} \label{subsec:evaluating}



\subsubsection*{Evaluating Experiment 1 and Experiment 2.}
In the first and second experiments, we assess the sensitivity of our approach by measuring top-$1$ and top-$5$ accuracy.
This is because for our approach to be useful to a game developer, the search system should be able to reliably identify objects specified in the text queries.
Top-$k$ accuracy is a binary measure; if there is a correct result in the top-$k$ results, the accuracy is $100\%$, otherwise the accuracy is $0\%$ -- there are no possible values in between. 



\subsubsection*{Evaluating Experiment 3}
In the third experiment, we measured the accuracy of our approach using recall~@5.
The reason for this choice is that we want to see what proportion of videos are relevant to the bug query, and how susceptible our system is to false positives when searching with bug queries. 
It is possible to report recall at higher levels, but the problem is that we cannot know how many videos in the dataset exactly match the search query without manually checking every video.
Recall~@5 is $100\%$ when all five out of five retrieved videos match the bug query, $80\%$ when four out of five retrieved videos match, etc. until $0\%$ when there are no matching videos.

 	\section{Results} \label{sec:results}
In this section, we present the results of the three experiments we designed to examine the ability of our proposed search system. 


\begin{table}[t!]
\centering
\caption{Average top-$k$ accuracy (\%) per game for simple queries (Experiment~1).}
\label{tab:pergameresults1}
\begin{tabular*}{\linewidth}{l@{\extracolsep{\fill}}l@{\extracolsep{\fill}}r@{\extracolsep{\fill}}r@{\extracolsep{\fill}}r@{\extracolsep{\fill}}r@{\extracolsep{\fill}}r@{\extracolsep{\fill}}r@{\extracolsep{\fill}}r@{\extracolsep{\fill}}r}
\toprule
 {} & \textbf{} & \textbf{{\texttt{GTA}}} & \textbf{{\texttt{RDR}}} & \textbf{{\texttt{JC3}}} & \textbf{{\texttt{F4}}} & \textbf{{\texttt{FC5}}} & \textbf{{\texttt{C77}}} & \textbf{{\texttt{ESV}}} & \textbf{{\texttt{W3}}} \\
 \midrule
\textbf{ViT-B/32} & \textbf{Top-$1$} & 74 & 71 & 61 & 65 & 50 & 55 & 54 & 54 \\
 & \textbf{Top-$5$} & 89 & 86 & 67 & 71 & 88 & 73 & 62 & 62 \\
 \midrule
\textbf{RN101} & \textbf{Top-$1$} & 84 & 50 & 61 & 59 & 59 & 43 & 62 & 62 \\
 & \textbf{Top-$5$} & 89 & 79 & 83 & 82 & 94 & 71 & 92 & 85 \\
 \bottomrule
\end{tabular*}
\end{table}

\begin{table}[t!]
\centering
\caption{Average top-$k$ accuracy (\%) per game for compound queries (Experiment~2).}
\label{tab:pergameresults2}
\begin{tabular*}{\linewidth}{l@{\extracolsep{\fill}}l@{\extracolsep{\fill}}r@{\extracolsep{\fill}}r@{\extracolsep{\fill}}r@{\extracolsep{\fill}}r@{\extracolsep{\fill}}r@{\extracolsep{\fill}}r@{\extracolsep{\fill}}r@{\extracolsep{\fill}}r}
\toprule
{} & \textbf{} & \textbf{\texttt{GTA}} & \textbf{\texttt{RDR}} & \textbf{\texttt{JC3}} & \textbf{\texttt{F4}} & \textbf{\texttt{FC5}} & \textbf{\texttt{C77}} & \textbf{\texttt{ESV}} & \textbf{\texttt{W3}} \\
\midrule
\textbf{ViT-B/32} & \textbf{Top-$1$} & 68 & 88 & 56 & 43 & 31 & 50 & 56 & 56 \\
 & \textbf{Top-$5$} & 100 & 100 & 81 & 64 & 69 & 75 & 89 & 67 \\
\midrule
\textbf{RN101} & \textbf{Top-$1$} & 84 & 88 & 31 & 36 & 56 & 67 & 33 & 44 \\
 & \textbf{Top-$5$} & 95 & 100 & 75 & 79 & 94 & 83 & 78 & 56 \\        
 \bottomrule
\end{tabular*}
\end{table}

\begin{table}[t!]
    \caption{Average top-$k$ accuracy (\%) per query for simple queries (Experiment~1). $N$ is the number of games searched.}
    \label{tab:results_t1}
    \centering
    \begin{tabular*}{\linewidth}{l @{\extracolsep{\fill}} r r @{\extracolsep{\fill}} r r @{\extracolsep{\fill}} r}
    \toprule
    \textbf{} & \textbf{} & \multicolumn{2}{l}{\textbf{ViT-B/32}} & \multicolumn{2}{l}{\textbf{RN101}} \\
    \textbf{Query} & \textbf{$N$} & \textbf{Top-$1$} & \textbf{Top-$5$} & \textbf{Top-$1$} & \textbf{Top-$5$} \\
    \midrule
    \textit{Airplane} & 4 & 75 & 100 & 100 & 100 \\
    \textit{Bear} & 5 & 80 & 100 & 60 & 100 \\
    \textit{Bike} & 4 & 50 & 75 & 50 & 100 \\
    \textit{Bridge} & 8 & 88 & 88 & 50 & 100 \\
    \textit{Car} & 5 & 80 & 100 & 80 & 100 \\
    \textit{Carriage} & 4 & 50 & 50 & 75 & 100 \\
    \textit{Cat} & 6 & 33 & 50 & 33 & 67 \\
    \textit{Cow} & 8 & 63 & 75 & 25 & 75 \\
    \textit{Deer} & 7 & 57 & 71 & 75 & 100 \\
    \textit{Dog} & 8 & 25 & 38 & 38 & 63 \\
    \textit{Fire} & 8 & 88 & 100 & 100 & 100 \\
    \textit{Helicopter} & 5 & 60 & 60 & 60 & 100 \\
    \textit{Horse} & 3 & 67 & 100 & 100 & 100 \\
    \textit{Mountain} & 7 & 100 & 100 & 100 & 100 \\
    \textit{Parachute} & 2 & 0 & 67 & 67 & 100 \\
    \textit{Ship} & 8 & 50 & 63 & 38 & 75 \\
    \textit{Snow} & 6 & 67 & 83 & 33 & 50 \\
    \textit{Tank} & 3 & 67 & 67 & 100 & 100 \\
    \textit{Traffic Light} & 5 & 40 & 40 & 20 & 20 \\
    \textit{Train} & 5 & 80 & 100 & 17 & 67 \\
    \textit{Truck} & 4 & 75 & 100 & 100 & 100 \\
    \textit{Wolf} & 6 & 17 & 50 & 86 & 86 \\
    \midrule
    \textbf{Average} & \textbf{5.5} & \textbf{60} & \textbf{76} & \textbf{64} & \textbf{86} \\
    \bottomrule
    \end{tabular*}
\end{table}

\subsubsection*{Results for simple queries (Experiment~1)}
In the first experiment we measured the top-$1$ and top-$5$ accuracy of our system with simple queries. 
The average accuracy for experiment 1 per game can be seen in Table~\ref{tab:pergameresults1}, and per query in Table~\ref{tab:results_t1}.
The  overall average top-$1$ accuracy and average top-$5$ accuracy for `ViT-B/32' is 60\% and 76\% respectively, and for `RN101' we have 64\% and 86\% respectively.
These results show that our system can identify a majority of objects in the game without fine-tuning or re-training.

\subsubsection*{Results for compound queries (Experiment~2)}
In the second experiment we measure the top-$1$ and top-$5$ accuracy of our approach with compound queries.
The average accuracy for experiment 2 per game can be seen in Table~\ref{tab:pergameresults2}, and per query in Table~\ref{tab:results_t2}.
For the second experiment, we find that our approach shows particularly high performance for all of our selected games, except for The Witcher 3: Wild Hunt.
Our approach achieves an overall average top-$5$ accuracy of \textbf{78\%} using `ViT-B/32' and \textbf{82\%} using the `RN101' model.
These results show that our approach is flexible enough to effectively search gameplay videos with compound queries. 

\begin{table}[t!]
    \caption{Average top-$k$ accuracy (\%) per query for compound queries (Experiment~2). $N$ is the number of games searched.}
    \label{tab:results_t2}
    \centering
    \begin{tabular*}{\linewidth}{l @{\extracolsep{\fill}} r r @{\extracolsep{\fill}} r r @{\extracolsep{\fill}} r}
    \toprule
    \textbf{} & \textbf{} & \multicolumn{2}{l}{\textbf{ViT-B/32}} & \multicolumn{2}{l}{\textbf{RN101}} \\
    \textbf{Query} & \textbf{$N$} & \textbf{Top-$1$} & \textbf{Top-$5$} & \textbf{Top-$1$} & \textbf{Top-$5$} \\
    \midrule
    \textit{A bald person} & 8 & 75 & 88 & 88 & 88 \\
    \textit{A bike on a mountain} & 4 & 25 & 75 & 50 & 75 \\
    \textit{A black car} & 5 & 80 & 100 & 80 & 100 \\
    \textit{A blue airplane} & 4 & 25 & 75 & 50 & 75 \\
    \textit{A blue car} & 5 & 80 & 80 & 40 & 100 \\
    \textit{A brown cow} & 7 & 29 & 71 & 57 & 71 \\
    \textit{A brown horse} & 3 & 100 & 100 & 100 & 100 \\
    \textit{A car on a mountain} & 4 & 75 & 75 & 75 & 100 \\
    \textit{A golden dragon} & 2 & 0 & 50 & 0 & 50 \\
    \textit{A gray tank} & 3 & 33 & 67 & 33 & 33 \\
    \textit{A man on top of a tank} & 4 & 25 & 50 & 0 & 0 \\
    \textit{A person in a jungle} & 7 & 57 & 100 & 57 & 100 \\
    \textit{A person on a mountain} & 7 & 71 & 100 & 57 & 100 \\
    \textit{A person wearing gold} & 8 & 50 & 88 & 50 & 100 \\
    \textit{A person wearing purple} & 8 & 50 & 88 & 25 & 63 \\
    \textit{A person with a pig mask} & 1 & 100 & 100 & 100 & 100 \\
    \textit{A police car} & 3 & 33 & 67 & 67 & 100 \\
    \textit{A police officer} & 3 & 33 & 33 & 67 & 100 \\
    \textit{A red car} & 5 & 80 & 100 & 80 & 100 \\
    \textit{A white airplane} & 4 & 75 & 75 & 50 & 100 \\
    \textit{A white horse} & 3 & 33 & 67 & 33 & 67 \\
    \textit{A white truck} & 5 & 40 & 60 & 60 & 80 \\
    \midrule
    \textbf{Average} & \textbf{4.7} & \textbf{53} & \textbf{78} & \textbf{55} & \textbf{82} \\
    \bottomrule
    \end{tabular*}
\end{table}

\begin{table*}[t!]
     \caption{Recall @5 (\%) for bug queries (Experiment~3). Queries that were not used per game are shown with values of `-'.}
     \label{tab:results_t3}
     \centering
\begin{tabular*}{\textwidth}{lr@{\extracolsep{\fill}}r@{\extracolsep{\fill}}r@{\extracolsep{\fill}}r@{\extracolsep{\fill}}r@{\extracolsep{\fill}}r@{\extracolsep{\fill}}r@{\extracolsep{\fill}}rr@{\extracolsep{\fill}}r@{\extracolsep{\fill}}r@{\extracolsep{\fill}}r@{\extracolsep{\fill}}r@{\extracolsep{\fill}}r@{\extracolsep{\fill}}r@{\extracolsep{\fill}}r}
    \toprule
     & \multicolumn{8}{l}{\textbf{ViT-B/32}} & \multicolumn{8}{l}{\textbf{RN101}} \\
     \textbf{Query}  & \smaller\textbf{\texttt{GTA}}  & \smaller\textbf{\texttt{RDR}} & \smaller\textbf{\texttt{JC3}}  & \smaller\textbf{\texttt{F4}} & \smaller\textbf{\texttt{ESV}} & \smaller\textbf{\texttt{W3}}& \smaller\textbf{\texttt{C77}} & \smaller\textbf{\texttt{FC5}} & \smaller\textbf{\texttt{GTA}} & \smaller\textbf{\texttt{RDR}} & \smaller\textbf{\texttt{JC3}} & \smaller\textbf{\texttt{F4}} & \smaller\textbf{\texttt{ESV}}  & \smaller\textbf{\texttt{W3}}& \smaller\textbf{\texttt{C77}} & \smaller\textbf{\texttt{FC5}} \\
    \midrule
\smaller \textit{A bike inside a car}                & \smaller 40  & \smaller - & \smaller - & \smaller - & \smaller - & \smaller - & \smaller - & \smaller - & \smaller 20  & \smaller - & \smaller - & \smaller - & \smaller - & \smaller - & \smaller - & \smaller - \\
\smaller \textit{A bike on a wall}                   & \smaller 100 & \smaller - & \smaller - & \smaller - & \smaller - & \smaller - & \smaller - & \smaller - & \smaller 100 & \smaller - & \smaller - & \smaller - & \smaller - & \smaller - & \smaller - & \smaller - \\
\smaller \textit{A car flying in the air}            & \smaller 100 & \smaller - & \smaller 100 & \smaller 40  & \smaller - & \smaller - & \smaller 60  & \smaller 80  & \smaller 100 & \smaller - & \smaller 100 & \smaller 40  & \smaller - & \smaller - & \smaller 100 & \smaller 80  \\
\smaller \textit{A car on fire}                      & \smaller 60  & \smaller - & \smaller 80  & \smaller - & \smaller - & \smaller - & \smaller 60  & \smaller 80  & \smaller 60  & \smaller - & \smaller 100 & \smaller - & \smaller - & \smaller - & \smaller 80  & \smaller 100 \\
\smaller \textit{A car in vertical position}         & \smaller 100 & \smaller - & \smaller 100 & \smaller - & \smaller - & \smaller - & \smaller 80  & \smaller 100 & \smaller 100 & \smaller - & \smaller 100 & \smaller - & \smaller - & \smaller - & \smaller 80  & \smaller 60  \\
\smaller \textit{A car stuck in a rock}              & \smaller - & \smaller - & \smaller - & \smaller - & \smaller - & \smaller - & \smaller 40  & \smaller - & \smaller - & \smaller - & \smaller - & \smaller - & \smaller - & \smaller - & \smaller 20  & \smaller - \\
\smaller \textit{A car stuck in a tree}              & \smaller 60  & \smaller - & \smaller 40  & \smaller - & \smaller - & \smaller - & \smaller - & \smaller 60  & \smaller 100 & \smaller - & \smaller 60  & \smaller - & \smaller - & \smaller - & \smaller - & \smaller 40  \\
\smaller \textit{A car under ground}                 & \smaller - & \smaller - & \smaller - & \smaller - & \smaller - & \smaller - & \smaller 60  & \smaller - & \smaller - & \smaller - & \smaller - & \smaller - & \smaller - & \smaller - & \smaller 20  & \smaller - \\
\smaller \textit{A carriage running over a person}   & \smaller - & \smaller - & \smaller - & \smaller - & \smaller 20  & \smaller - & \smaller - & \smaller - & \smaller - & \smaller - & \smaller - & \smaller - & \smaller 40  & \smaller - & \smaller - & \smaller - \\
\smaller \textit{A dragon inside the floor}          & \smaller - & \smaller - & \smaller - & \smaller - & \smaller 20  & \smaller - & \smaller - & \smaller - & \smaller - & \smaller - & \smaller - & \smaller - & \smaller 60  & \smaller - & \smaller - & \smaller - \\
\smaller \textit{A head without a body}              & \smaller - & \smaller - & \smaller - & \smaller 20  & \smaller - & \smaller - & \smaller - & \smaller - & \smaller - & \smaller - & \smaller - & \smaller 0   & \smaller - & \smaller - & \smaller - & \smaller - \\
\smaller \textit{A headless person}                  & \smaller - & \smaller 20  & \smaller - & \smaller - & \smaller - & \smaller - & \smaller - & \smaller - & \smaller - & \smaller 20  & \smaller - & \smaller - & \smaller - & \smaller - & \smaller - & \smaller - \\
\smaller \textit{A helicopter inside a car}          & \smaller - & \smaller - & \smaller - & \smaller - & \smaller - & \smaller - & \smaller - & \smaller 20  & \smaller - & \smaller - & \smaller - & \smaller - & \smaller - & \smaller - & \smaller - & \smaller 40  \\
\smaller \textit{A horse floating the air}           & \smaller - & \smaller - & \smaller - & \smaller - & \smaller 100 & \smaller - & \smaller - & \smaller - & \smaller - & \smaller - & \smaller - & \smaller - & \smaller 100 & \smaller - & \smaller - & \smaller - \\
\smaller \textit{A horse in the air}                 & \smaller - & \smaller 80  & \smaller - & \smaller - & \smaller - & \smaller 100 & \smaller - & \smaller - & \smaller - & \smaller 100 & \smaller - & \smaller - & \smaller - & \smaller 100 & \smaller - & \smaller - \\
\smaller \textit{A horse in the fire}                & \smaller - & \smaller 40  & \smaller - & \smaller - & \smaller - & \smaller 20  & \smaller - & \smaller - & \smaller - & \smaller 20  & \smaller - & \smaller - & \smaller - & \smaller 20  & \smaller - & \smaller - \\
\smaller \textit{A horse on fire}                    & \smaller - & \smaller - & \smaller - & \smaller - & \smaller 20  & \smaller - & \smaller - & \smaller - & \smaller - & \smaller - & \smaller - & \smaller - & \smaller 20  & \smaller - & \smaller - & \smaller - \\
\smaller \textit{A horse on top of a building}       & \smaller - & \smaller 60  & \smaller - & \smaller - & \smaller - & \smaller - & \smaller - & \smaller - & \smaller - & \smaller 20  & \smaller - & \smaller - & \smaller - & \smaller - & \smaller - & \smaller - \\
\smaller \textit{A horse to stand on its legs}       & \smaller - & \smaller - & \smaller - & \smaller - & \smaller - & \smaller 60  & \smaller - & \smaller - & \smaller - & \smaller - & \smaller - & \smaller - & \smaller - & \smaller 100 & \smaller - & \smaller - \\
\smaller \textit{A person falling inside the ground} & \smaller - & \smaller - & \smaller - & \smaller 20  & \smaller - & \smaller - & \smaller - & \smaller - & \smaller - & \smaller - & \smaller - & \smaller 40  & \smaller - & \smaller - & \smaller - & \smaller - \\
\smaller \textit{A person flying in the air}         & \smaller 80  & \smaller 100 & \smaller 100 & \smaller 60  & \smaller 40  & \smaller 100 & \smaller 80  & \smaller 100 & \smaller 100 & \smaller 100 & \smaller 80  & \smaller 100 & \smaller 60  & \smaller 100 & \smaller 80  & \smaller 100 \\
\smaller \textit{A person goes through the ground}   & \smaller - & \smaller 40  & \smaller - & \smaller - & \smaller - & \smaller - & \smaller - & \smaller - & \smaller - & \smaller 0   & \smaller - & \smaller - & \smaller - & \smaller - & \smaller - & \smaller - \\
\smaller \textit{A person in fire}                   & \smaller - & \smaller 100 & \smaller - & \smaller 60  & \smaller 60  & \smaller 100 & \smaller - & \smaller - & \smaller - & \smaller 100 & \smaller - & \smaller 80  & \smaller 60  & \smaller 80  & \smaller - & \smaller - \\
\smaller \textit{A person inside a chair}            & \smaller - & \smaller - & \smaller - & \smaller 100 & \smaller 40  & \smaller - & \smaller - & \smaller - & \smaller - & \smaller - & \smaller - & \smaller 40  & \smaller 40  & \smaller - & \smaller - & \smaller - \\
\smaller \textit{A person inside a rock}             & \smaller - & \smaller - & \smaller - & \smaller - & \smaller - & \smaller 80  & \smaller - & \smaller - & \smaller - & \smaller - & \smaller - & \smaller - & \smaller - & \smaller 40  & \smaller - & \smaller - \\
\smaller \textit{A person on the house wall}         & \smaller - & \smaller - & \smaller - & \smaller - & \smaller - & \smaller 60  & \smaller - & \smaller - & \smaller - & \smaller - & \smaller - & \smaller - & \smaller - & \smaller 40  & \smaller - & \smaller - \\
\smaller \textit{A person stuck in a barrel}         & \smaller - & \smaller - & \smaller - & \smaller - & \smaller 60  & \smaller - & \smaller - & \smaller - & \smaller - & \smaller - & \smaller - & \smaller - & \smaller 40  & \smaller - & \smaller - & \smaller - \\
\smaller \textit{A person stuck in a tree}           & \smaller 80  & \smaller - & \smaller - & \smaller - & \smaller - & \smaller 40  & \smaller - & \smaller - & \smaller 80  & \smaller - & \smaller - & \smaller - & \smaller - & \smaller 40  & \smaller - & \smaller - \\
\smaller \textit{A person stuck inside a wall}       & \smaller - & \smaller - & \smaller - & \smaller 20  & \smaller - & \smaller - & \smaller - & \smaller - & \smaller - & \smaller - & \smaller - & \smaller 40  & \smaller - & \smaller - & \smaller - & \smaller - \\
\smaller \textit{A person stuck on the }ceiling      & \smaller - & \smaller - & \smaller - & \smaller - & \smaller 40  & \smaller - & \smaller - & \smaller - & \smaller - & \smaller - & \smaller - & \smaller - & \smaller 40  & \smaller - & \smaller - & \smaller - \\
\smaller \textit{A person under a  vehicle}          & \smaller 80  & \smaller - & \smaller - & \smaller 60  & \smaller - & \smaller - & \smaller 20  & \smaller - & \smaller 60  & \smaller - & \smaller - & \smaller 60  & \smaller - & \smaller - & \smaller 0   & \smaller - \\
\smaller \textit{A person under a car}               & \smaller 60  & \smaller - & \smaller - & \smaller - & \smaller - & \smaller - & \smaller - & \smaller - & \smaller 60  & \smaller - & \smaller - & \smaller - & \smaller - & \smaller - & \smaller - & \smaller - \\
\smaller \textit{A person under a vehicle}           & \smaller - & \smaller - & \smaller - & \smaller - & \smaller - & \smaller - & \smaller - & \smaller 60  & \smaller - & \smaller - & \smaller - & \smaller - & \smaller - & \smaller - & \smaller - & \smaller 80  \\
\smaller \textit{A person under the carriage}        & \smaller - & \smaller 40  & \smaller - & \smaller - & \smaller - & \smaller - & \smaller - & \smaller - & \smaller - & \smaller 40  & \smaller - & \smaller - & \smaller - & \smaller - & \smaller - & \smaller - \\
\smaller \textit{A person without head}              & \smaller - & \smaller - & \smaller - & \smaller 20  & \smaller - & \smaller - & \smaller - & \smaller - & \smaller - & \smaller - & \smaller - & \smaller 20  & \smaller - & \smaller - & \smaller - & \smaller - \\
\smaller \textit{A ship under water}                 & \smaller - & \smaller - & \smaller - & \smaller - & \smaller - & \smaller 40  & \smaller - & \smaller - & \smaller - & \smaller - & \smaller - & \smaller - & \smaller - & \smaller 80  & \smaller - & \smaller - \\
\smaller \textit{A tank in the air}                  & \smaller 80  & \smaller - & \smaller 100 & \smaller - & \smaller - & \smaller - & \smaller - & \smaller - & \smaller 80  & \smaller - & \smaller 80  & \smaller - & \smaller - & \smaller - & \smaller - & \smaller - \\
\smaller \textit{A vehicle inside the water}         & \smaller 80  & \smaller - & \smaller 80  & \smaller 40  & \smaller - & \smaller - & \smaller 80  & \smaller 100 & \smaller 80  & \smaller - & \smaller 100 & \smaller 40  & \smaller - & \smaller - & \smaller 40  & \smaller 80  \\
\smaller \textit{A vehicle on top of building}       & \smaller 100 & \smaller - & \smaller 100 & \smaller - & \smaller - & \smaller - & \smaller 100 & \smaller - & \smaller 100 & \smaller - & \smaller 100 & \smaller - & \smaller - & \smaller - & \smaller 100 & \smaller - \\
\smaller \textit{A vehicle on top of rooftop}        & \smaller 60  & \smaller - & \smaller 80  & \smaller - & \smaller - & \smaller - & \smaller - & \smaller - & \smaller 100 & \smaller - & \smaller 80  & \smaller - & \smaller - & \smaller - & \smaller - & \smaller - \\
\smaller  \textit{An airplane in a tree}              & \smaller - & \smaller - & \smaller - & \smaller - & \smaller - & \smaller - & \smaller - & \smaller 100 & \smaller - & \smaller - & \smaller - & \smaller - & \smaller - & \smaller - & \smaller - & \smaller 80  \\
\smaller  \textit{An airplane in the water}           & \smaller 20  & \smaller - & \smaller 60 & \smaller - & \smaller - & \smaller - & \smaller - & \smaller 60  & \smaller 40  & \smaller - & \smaller 80 & \smaller - & \smaller - & \smaller - & \smaller - & \smaller 80  \\
\smaller  \textit{Cars in accident}                   & \smaller 100 & \smaller - & \smaller 60  & \smaller - & \smaller - & \smaller - & \smaller 100 & \smaller 100 & \smaller 80  & \smaller - & \smaller 80  & \smaller - & \smaller - & \smaller - & \smaller 100 & \smaller 100 \\
\smaller  \textit{Two cars on top of each other}       & \smaller - & \smaller - & \smaller - & \smaller - & \smaller - & \smaller - & \smaller 60  & \smaller - & \smaller - & \smaller - & \smaller - & \smaller - & \smaller - & \smaller - & \smaller 40  & \smaller -   \\          
    \midrule
    \smaller\textbf{Average} & \smaller\textbf{77}	& \smaller\textbf{60}	& \smaller\textbf{82}	& \smaller\textbf{44}	& \smaller\textbf{44}	& \smaller\textbf{67}	& \smaller\textbf{67}	& \smaller\textbf{78}	& \smaller\textbf{83}	& \smaller\textbf{50}	& \smaller\textbf{87}	& \smaller\textbf{46}	& \smaller\textbf{51}	& \smaller\textbf{67}	& \smaller\textbf{60}	& \smaller\textbf{76} \\
    \bottomrule
\end{tabular*}
\end{table*}

\subsubsection*{Results for bug queries (Experiment~3)}
In the final experiment, we measure recall~@5 of our approach with bug queries. 
Table~\ref{tab:results_t3} shows the results for Experiment~3 for each query with each game.
Our approach shows particularly high performance for Grand Theft Auto~V, Just Cause~3, and Far Cry~5.
The average accuracy for Experiment~3 across all 44 unique queries is 66.12\% and 66.35\% using `ViT-B/32' and `RN101' respectively. 
These numbers suggest that, in most cases, our approach can reliably retrieve relevant videos based on an English text query containing a description of an event. 
Moreover, we can conclude that contrastive pre-training methods are powerful enough to be used in the video game domain, especially for bug detection in gameplay videos. 

 	\section{Discussion} \label{sec:discussion}

\begin{figure*}[!t]
	\centering
	\begin{subfigure}[t]{\columnwidth}
		\centering
		\includegraphics[width=\textwidth]{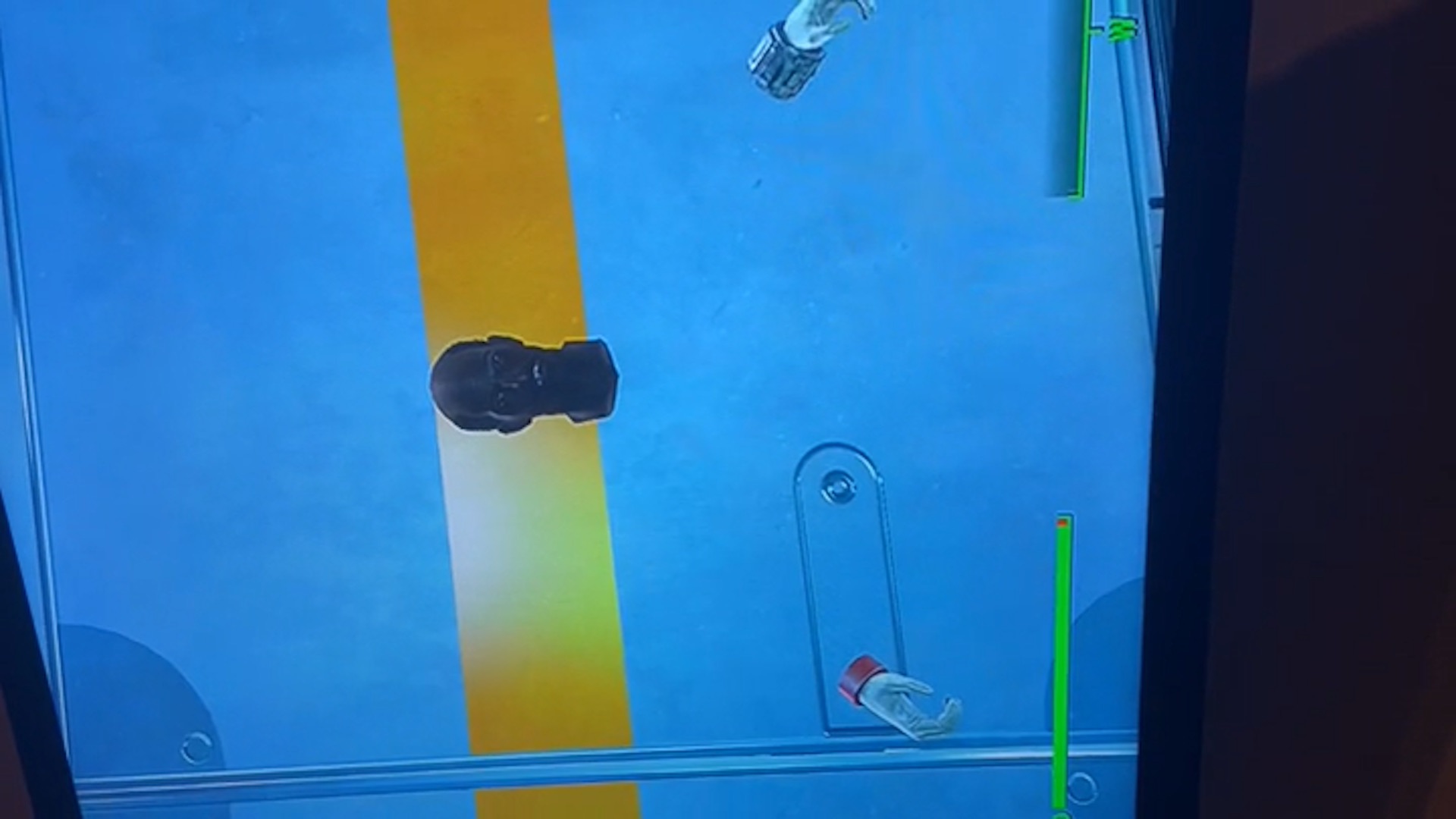}
		\caption{\reddit{kl5d3z}{Video} of \textit{`A head without a body'} from Fallout~4.}
		\Description{Fully described in the text.}
		\label{fig:sample_pair_1}
	\end{subfigure}
	\hfill
	\begin{subfigure}[t]{\columnwidth}
		\centering
		\includegraphics[width=\textwidth]{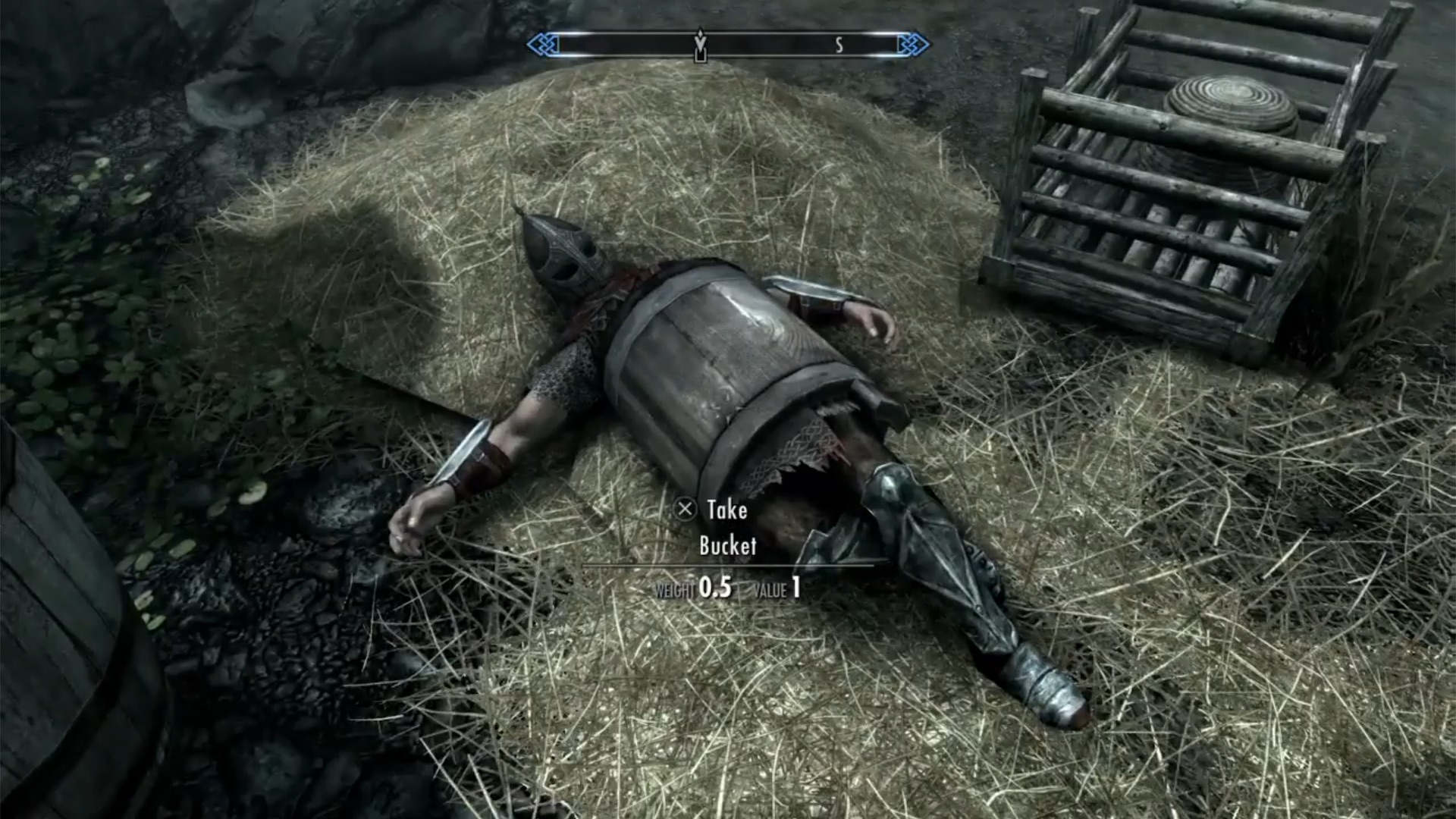}
		\caption{\reddit{g5pm35}{Video} of \textit{`A person stuck in a barrel'} from The Elder Scrolls~V: Skyrim.}
		\Description{Fully described in the text.}
		\label{fig:sample_pair_2}
	\end{subfigure}
	\hfill
	\begin{subfigure}[t]{\columnwidth}
		\centering
		\includegraphics[width=\textwidth]{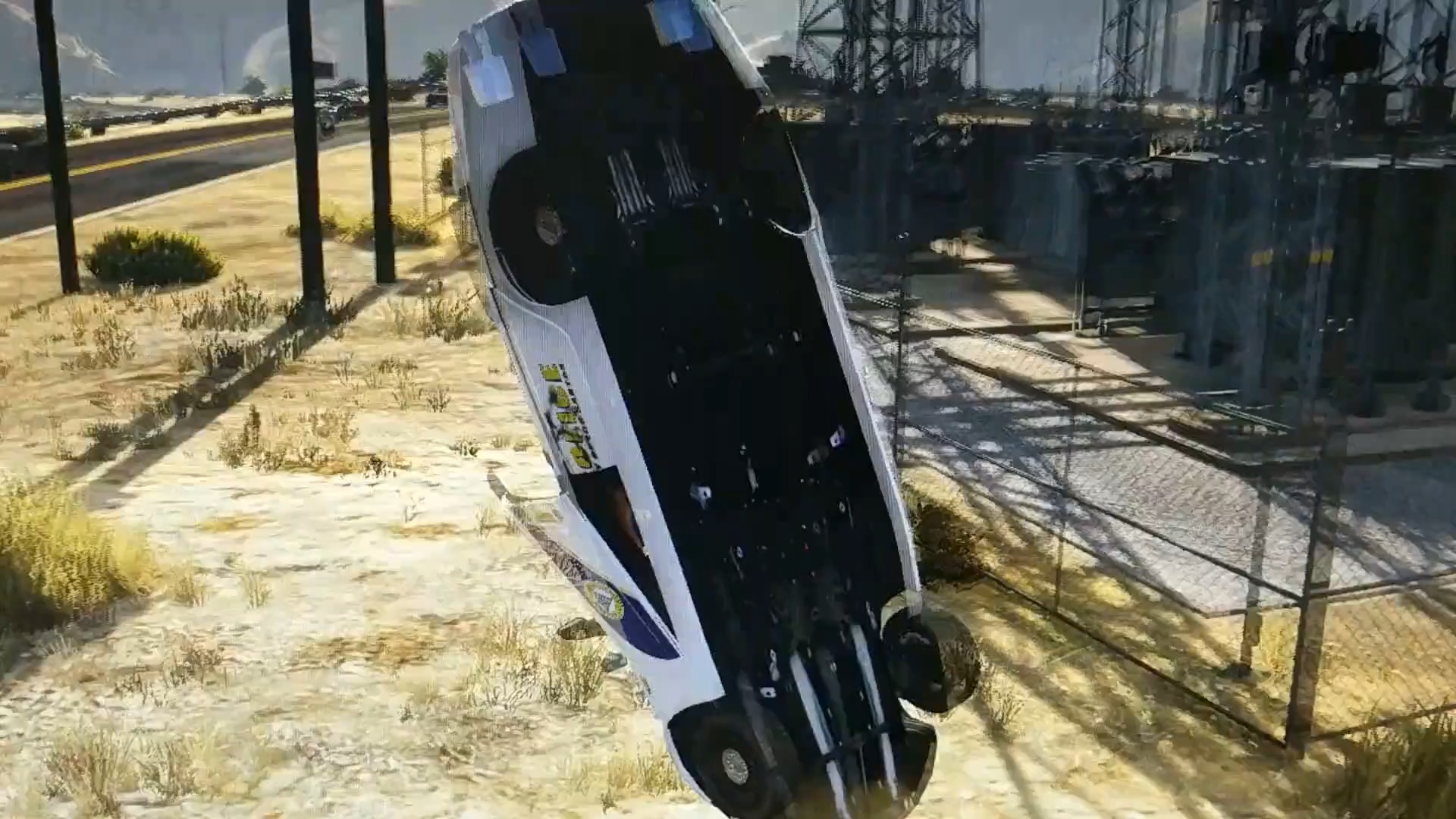}
		\caption{\reddit{6xplqg}{Video} of \textit{`A car in a vertical position'} from Grand Theft Auto~V.}
		\Description{Fully described in the text.}
		\label{fig:sample_pair_3}
	\end{subfigure}
	\hfill
	\begin{subfigure}[t]{\columnwidth}
		\centering
		\includegraphics[width=\textwidth]{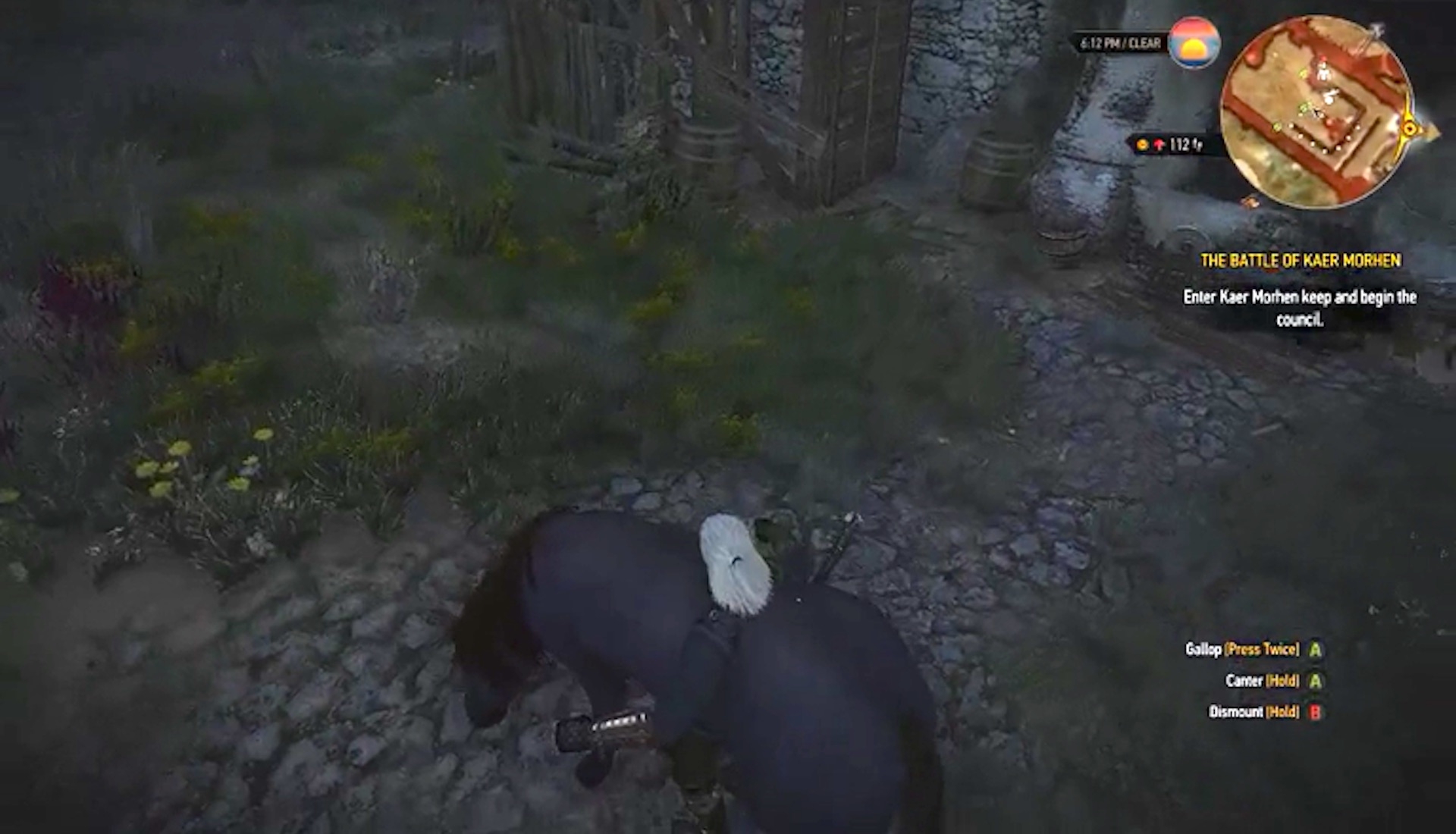}
		\caption{\reddit{8t7qfa}{Video} of \textit{`A person stuck in a horse'} from The Witcher~3: Wild Hunt.}
		\Description{Fully described in the text.}
		\label{fig:sample_pair_4}
	\end{subfigure}
	\caption{Relevant gameplay videos identified using our approach with bug queries.}
	\Description{Fully described in the text.}
	\label{fig:more_example_pairs}
\end{figure*}

In this section, we discuss the strengths and weaknesses of our approach, based on the results of our experiments.
Figure~\ref{fig:more_example_pairs} shows several example video frames from videos identified when searching gameplay videos with text queries using our approach.
These examples help to illustrate the promising potential of our approach. 
Given that our method does not require any training on gameplay videos, our zero-shot object and event identification results are promising.
During our experiments, the first author manually analyzed each video returned by our search approach, including false positives.
Below, the causes of false positives in our search results are detailed.

\subsubsection*{Adversarial poses}
One important category of problems is the unusual pose of familiar objects. As extensively tested and reported by Alcorn et al. \cite{alcorn2019strike}, neural networks occasionally misclassify objects when they have different poses than what they used to have in the training set. For example, consider a neural network that can detect a `car' in an image. It is possible to find a particular camera angle for which the neural network can not detect the `car' in that image. In a dataset of natural images, we might have lots of cars, but the camera angle and the position of the car relative to the camera do not vary a lot. A neural network trained on these datasets will struggle to detect a car when it sees it from a very unusual angle (e.g., when it is positioned vertically)

\subsubsection*{Confusing textures and patterns in the images}
The textures and patterns can pose influential distractions and confusion for the neural network model in our approach. 
Sometimes a part of a game environment has a texture similar to another object. 
For example, our model confuses a striped colored wall in the Grand Theft Auto~V game with a `parachute.'  
This category of problems is hard to encounter globally because each game has a diverse look and feel and creative artistic directions.

\subsubsection*{Confusion about different types of vehicles}
During analysis of the videos that contain bugs related to cars, we noticed that sometimes some of the results partially match the textual description, except we see a bike instead of a car. 
Through our manual evaluation of the search results, we found that the model in our approach sometimes confused cars, bikes, airplanes, tanks, and other military vehicles. 
An instance of this misclassification is when we search for `A car on fire'. 
In some of the retrieved videos, we saw an airplane on fire instead of a car. 

\subsubsection*{Confusion about different four-legged animals}
After reviewing several results for queries related to animals, we found out that the model in our approach struggles to distinguish different animals. 
More specifically, for gameplay videos the CLIP model will confuse `dogs', `cats', `deer', `wolves', and sometimes `cows' and 'horses' with each other. 
A possible remedy for this problem is getting help from a separate pre-trained animal detection model to verify the CLIP model's prediction. 

 	\section{Limitations} \label{sec:limitations}

\subsection{Adversarial samples} \label{subsec:adversarial}
Every machine learning method suffers from a group of adversarial attacks and out-of-distribution samples. As described extensively in previous work~\cite{alcorn2019strike}, any data point outside the training distribution is problematic for machine learning algorithms.
Similarly, we observe some cases in which the neural network model makes an incorrect prediction. In particular, our model has some difficulty making a correct guess if it saw an object in an unfamiliar or adversarial pose. Due to physical simulations in video games, these adversarial poses are prevalent.

Another observation we had is about text patches inside the games. The CLIP model has the ability to `read' the text inside an image as well. This feature is not something that the model was explicitly trained for, but rather some emergent behavior of pre-training in a contrastive setting. Sometimes searching a particular text query will result in retrieving the video that ignores the meaning of the text query, but the image contains that text. For example, if any video frames include a text field containing `a blue car', searching for the query `a blue car', will retrieve that video. Obviously, depending on the use case, this can be treated as both a feature and bug.


\subsection{Improvements on search speed} \label{subsec:searchspeed}
In our proposed method, we calculate the embeddings of all frames in advance. With this pre-processing step, our system answers an arbitrary text query in just a few seconds. It might not be possible to perform this step in advance for some exceptional use cases. For handling such cases, there are some performance improvement techniques to run each neural network faster in inference mode, at the cost of sacrificing the model's accuracy. For example, it is possible to reduce the floating point precision of a model \cite{courbariaux2014training} or even binarize the entire model \cite{hubara2016binarized}. One simple but effective way to achieve faster runtime is to cut the last layers of the neural network gradually to reach an optimal performance vs. accuracy trade-off \cite{9474052}. 
Using these techniques, or similar speed-up approaches, improving the presented system is possible.

 	\section{Threats to validity} \label{sec:threats}

\subsubsection*{Threats to internal validity.}
Due to a lack of a benchmark dataset, we designed a set of custom queries for searching the gameplay videos in our \texttt{GamePhysics} dataset.
To address potential bias when generating these queries, the first author performed a pilot analysis of 80 gameplay videos across the 8 selected games to determine relevant objects and events before we designed the queries.

In each of our experiments, we assumed that an accuracy measurement of 0\% indicated that our approach failed to correctly identify any relevant videos.
For example, in Experiment~3 we assumed that a recall~@5 of $0\%$ in our search results indicated that our approach failed to identify that bug query in that game.
However, it could instead be the case that our dataset does not contain any videos that match the query.
Without a benchmark dataset, we do not have the ground truth for whether a repository of gameplay videos contains any matches for a given arbitrary text query.
This means that the reported performance values are possibly lower estimates of the actual performance.

In Experiment~3, we used our second aggregation method (Section~\ref{subsec:aggregating}), which involved the selection of a pool size hyperparameter. 
Although we selected the default value of 1,000 based on manual trial and error, different selections of this hyperparameter could lead to different results for Experiment~3.
Therefore, future research is required to understand how the selection of the pool size in our second aggregation method impacts the performance of our approach.

\subsubsection*{Threats to external validity.}
While our dataset predominantly consists of gameplay videos that contain game physics bugs, our approach may not be as effective with other datasets of gameplay videos.
Non-curated datasets may contain many more false positives (non-buggy gameplay), for example if using gameplay streaming footage.
Additionally, we excluded long ($>$60 seconds) videos, meaning our approach may not be effective for long videos. We also ignored all videos with scores of zero from the GamePhysics subreddit. After manually checking a random sample of low-scored posts we observed that a score of 0 almost always indicated low quality/spam/etc.  This threshold might not be generalizable to other subreddits. 
Future research is required to evaluate the performance of our approach with long videos and non-curated datasets.

 	\section{Conclusion} \label{sec:conclusion}
In this paper, we proposed a novel approach to mine large repositories of gameplay videos by leveraging the zero-shot transfer capabilities of CLIP to connect video frames with an English text query.
Our approach is capable of finding objects in a large dataset of videos, using simple and compound queries. 
Additionally, our approach shows promising performance in finding specific (bug-related) events, indicating it has the potential to be applied in automatic bug identification for video games.
Even though we did not perform any fine-tuning or re-training to adapt the CLIP model to the video game domain, our approach performs surprisingly well on the majority of video games. 
We evaluated our system on a dataset of \textbf{6,192} videos from eight games with different visual styles and elements. 
When experimenting with the bug queries, we measured recall~@5 and found the average accuracy of our approach across all 44 unique bug queries is \textbf{66.24\%} when averaged across both of the CLIP architectures utilized in our experiments.
Furthermore, our manual analysis of the search results enabled us to discuss causes of false-positives in our approach and identify several future research directions. 
Our approach lays the foundation to utilizing contrastive learning models for zero-shot bug identification in video games.
Future work in this line of research will provide more insights into video games bugs, and will pave the way to creating a new paradigm of automated bug detection methods for video games.

 	
 	\AtNextBibliography{\footnotesize}
	\printbibliography
\end{document}